\documentclass[journal]{IEEEtran}

\ifCLASSINFOpdf
\else
\fi

\usepackage[ruled, linesnumbered]{algorithm2e}
\usepackage{algorithmicx}
\usepackage{algpseudocode}
\usepackage{color}
\usepackage{caption}
\usepackage{pifont}
\usepackage{times}
\usepackage{epsfig}
\usepackage{graphicx}
\usepackage{amsmath}
\usepackage{amssymb}
\usepackage{yhmath}
\usepackage{epsfig}
\usepackage{graphicx}
\usepackage{amsmath}
\usepackage{amssymb}
\usepackage{bm}
\usepackage{algorithm2e}
\usepackage{multirow}
\usepackage{subfigure}
\usepackage{amsthm}
\usepackage{booktabs}
\usepackage{tabularx}
\usepackage{url}
\usepackage{graphicx}
\usepackage{subfigure} 
\usepackage{mathrsfs}
\usepackage{amsmath}
\usepackage{multirow}
\usepackage{rotfloat}
\usepackage{booktabs}
\usepackage{graphicx}
\usepackage{bbding}

\usepackage{makecell}

\usepackage{booktabs}
\usepackage{bbding}
\usepackage{multirow}
\usepackage{pifont}

\usepackage{bm}
\usepackage{booktabs}
\usepackage{color}
\usepackage{rotfloat}
\usepackage{tabularx}
\usepackage[misc]{ifsym}

\usepackage{multirow}
\usepackage{float}

\usepackage{rotfloat}
\usepackage{tabularx}
\usepackage[misc]{ifsym}

\usepackage{multirow}
\usepackage{float}

\usepackage{graphicx}
\usepackage{makecell}

\usepackage{arydshln}

\begin{document}

\title{Learning Heavily-Degraded Prior for Underwater Object Detection}

\author{ Chenping Fu, Xin Fan, \textit{Senior Member, IEEE}, Jiewen Xiao, Wanqi Yuan, Risheng Liu, \textit{Member, IEEE},and Zhongxuan Luo
	\thanks{
		
		Xin Fan, Risheng Liu, and Zhongxuan Luo are with the DUT-RU
		International School of Information Science, Dalian University of Technology,
		Dalian 116024, China (Corresponding author: Xin Fan. e-mail: xin.fan@ieee.org)
		Chenping Fu, Jiewen Xiao, and Wanqi Yuan are with Dalian University of Technology, Dalian 116024, China. 
}	  
}

\markboth{Journal of \LaTeX\ Class Files,~Vol.~14, No.~8, August~2015}%
{Shell \MakeLowercase{\textit{et al.}}: Bare Demo of IEEEtran.cls for IEEE Journals}

\maketitle

\begin{abstract}
Underwater object detection suffers from low detection performance because the distance and wavelength dependent imaging process yield evident image quality degradations such as haze-like effects, low visibility, and color distortions. Therefore, we commit to resolving the issue of underwater object detection with compounded environmental degradations. Typical approaches attempt to develop sophisticated deep architecture to generate high-quality images or features. However, these methods are only work for limited ranges because imaging factors are either unstable, too sensitive, or compounded. Unlike these approaches catering for high-quality images or features, this paper seeks transferable prior knowledge from detector-friendly images. The prior guides detectors removing degradations that interfere with detection. It is based on statistical observations that, the heavily degraded regions of detector-friendly (DFUI) and underwater images have evident feature distribution gaps while the lightly degraded regions of them overlap each other. Therefore, we propose a residual feature transference module (RFTM) to learn a mapping between deep representations of the heavily degraded patches of DFUI- and underwater- images, and make the mapping as a heavily degraded prior (HDP) for underwater detection. Since the statistical properties are independent to image content, HDP can be learned without the supervision of semantic labels and plugged into popular CNN-based feature extraction networks to improve their performance on underwater object detection. Without bells and whistles, evaluations on URPC2020 and UODD show that our methods outperform CNN-based detectors by a large margin. Our method with higher speeds and less parameters still performs better than transformer-based detectors. Our code and DFUI dataset can be found in https://github.com/xiaoDetection/Learning-Heavily-Degraed-Prior.

\end{abstract}   
     

\IEEEpeerreviewmaketitle

\section{Introduction}

\IEEEPARstart{V}isual underwater object detection (UOD) has been playing an increasingly important role in fisheries, aquaculture, and marine resource investigation~\cite{houminjun,chenlong}. Underwater images exhibit significant appearance discrepancy with those captured in natural scenes owing to the complicated imaging process characterized as:
       
\begin{equation}
\label{eq:imaging}
\mathbf{I} = \mathbf{J}\cdot\mathbf{t}+ \mathbf{A}\cdot(\mathbf{1}-\mathbf{t}),
\end{equation}
where the operator $\cdot$ denotes the pixel-wise multiplication and $\mathbf{I}$, $\mathbf{J}$ and $\mathbf{A}$ are the observed image, scene radiance (ground truth), and atmospheric light, respectively. The transmission map $\mathbf{t}$ reflects the portion of scene radiance that reaches the imaging plane, scattered by particles in water. This distance and wavelength dependent transmission process yields evident quality degradations such as haze-like effects, low visibility, and color distortions. This paper addresses the issue of object detection with compounded environmental degradations that greatly challenges existing deep detectors gaining success in natural scenes~\cite{PDCNet,DRnet,ESCNet,RefineDet++}. 

Recently, researchers attempt to develop sophisticated deep architectures in order to enhance image quality or features favoring the UOD task. It is straightforward to cascade the network for underwater image enhancement (UIE) with a generic deep detector~\cite{FUNIE,jiang2022target,liu2022twin,MU22}. Unfortunately, the visually appealing output of the UIE module does~\emph{not} necessarily generate high accuracy for deep detectors~\cite{houminjun,chenlong}. Hence, researchers resort to a new architecture along with a designated loss for learning the features yielding better detection~\cite{chenlong,colorUIE}. Typically, these networks enhancing either quality or features are huge and inefficient for inference, hindering their applications to time critical scenarios.      

One alternative treatment for UOD is to re-train popular detectors \cite{Cascade,YOLOX,YOLOS,SWIN,detetcors,FoveaBox,DETR,grid} by collecting a large number of underwater training examples~\cite{UODD,FERNet,UDD}. However, they suffer a poor detection performance, since the environmental degradation masks many valuable features of a scene. There also exist works to synthesize underwater images from natural ones using generative adversarial networks~\cite{watergan,UEGAN}. These synthesized methods augment the quantity of training examples and introduce natural images guidance, but the effectiveness for real world UOD is questionable since underwater objects/scenes evidently differ from natural ones.


%
\begin{figure*}[!t]
	\centering
	\includegraphics[width=\textwidth,height=2.6in]{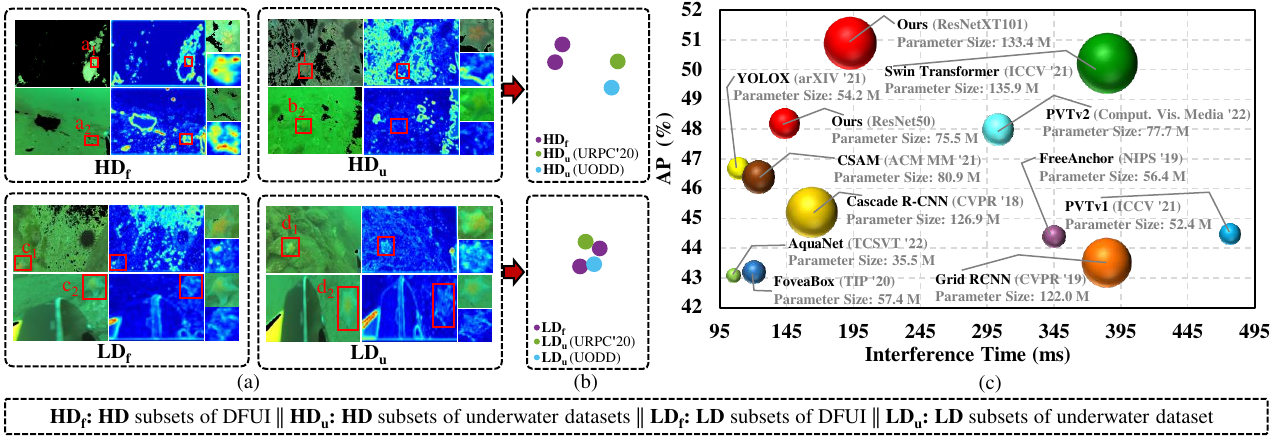}
	\caption{We choose starfish as examples, \textit{i.e.,} a$_{1,2}$, b$_{1,2}$, c$_{1,2}$, and d$_{1,2}$ from $\mathbf{HD_{f}}$, $\mathbf{HD}_{\mathbf{u}}$, $\mathbf{LD}_{\mathbf{f}}$, and $\mathbf{LD}_{\mathbf{u}}$, respectively. (a) the features visualization of these starfish using ResNet50 trained on DFUI. We can see that the feature response of $\mathbf{HD_{f}}$ and $\mathbf{HD}_{\mathbf{u}}$ is different while $\mathbf{LD}_{\mathbf{f}}$ and $\mathbf{LD}_{\mathbf{u}}$ is similar. (b) the t-SNE \cite{TSNE} diagram of the feature distribution. We can see that there is an evident feature distribution gap between $\mathbf{HD_{f}}$ and $\mathbf{HD}_{\mathbf{u}}$ while $\mathbf{LD}_{\mathbf{f}}$ and $\mathbf{LD}_{\mathbf{u}}$ overlap each other. (c) the accuracy-speed-size trade-off of accurate models on URPC'20.}
	\label{contri0}
\end{figure*}

This paper focuses on the heavily-degraded (HD) patches and bridges the distribution gap between these patches and those detector-friendly ones (DFUI). We pick the images, to which a detector~\cite{Cascade} applies and outputs high accuracy ($\mathbf{AP\ge60}$), from a large collection of underwater images, as the DFUI set. The patches with the transmission values $t$ less than a threshold from the DFUI and two publicly available datasets,~\emph{i.e.}, URPC2020~\footnote{http://www.cnurpc.org/} and UODD~\cite{UODD}, constitute the HD subsets of these three sets; those having higher transmission values produce the respective lightly degraded (LD) subsets. A typical example is shown in Fig.~\ref{contri0}(a)-(b). This plot reveals that the HD subset of the DFUI set distributes apart far from those of the other two while the LD subsets of all the three overlap each other. Thus, existing detectors fail learning favorable features from images heavily degraded by underwater environments and thus gain downgraded performance.


Specifically, we design a residual feature transference module (RFTM) to learn a mapping relationship between deep representations of the heavily-degraded regions (patches) of DFUI- and underwater- images, and make the mapping as a heavily degraded prior (HDP) for low-quality UOD task. The RFTM is devised to transfer the features of heavily degraded regions of image to those favoring object detectors, where a three layers forward network is designed to simulate residual transfer mechanism between different data distributions. This module can be plugged into the feature extraction network of any convolutional network based deep detectors, improving their respective performance. Fig.~\ref{contri0}(c) illustrates that our module embedded into two mainstream networks ResNeXT101 and ResNet50 gains superior accuracy over the latest CNN/Transformer-based detectors with less parameter size and inference time. We circumvent constructing and training complicated deep networks to enhance images or features as we only consider the HD images stumbling generic detectors. Meanwhile, the training of RFTM only demands unlabeled underwater image patches avoiding unrealistic synthesized examples that bring downside effects for real-world UOD. We summarize the contributions of this paper below:
\begin{itemize}
	\item This study discovers that the heavily-degraded regions of underwater and favoring object detector images have an evident gap, thus acting the bottleneck for real-world UOD tasks. As a nontrivial byproduct, we build a DFUI dataset with 5,265 images available for investigating the UOD task.
	\item We propose a residual feature transference module to learn the mapping between the heavily-degraded regions of the detector-friendly and generic underwater images. The module can be plugged into existing CNN-based feature extraction networks of object detectors, tackling the bottleneck for real-world UOD. 
	\item We devise a two-stage training scheme from the perspective of the unsupervised and finetune learning strategy yielding optimal network parameters for efficiently plugging RFTM into a detector.    
	\item Comprehensive experiments on the URPC2020 and UODD datasets show the superiority of our proposed methods. On the URPC2020 dataset, our module can increase $AP$ from 45.0 to 48.2 over the baseline ResNet50, and from 45.2 to 50.9 over ResNetXT101. Similar AP gains are also achieved on the UODD dataset. 
\end{itemize}

\begin{figure*}
	\centering
	\includegraphics[width=7.2in]{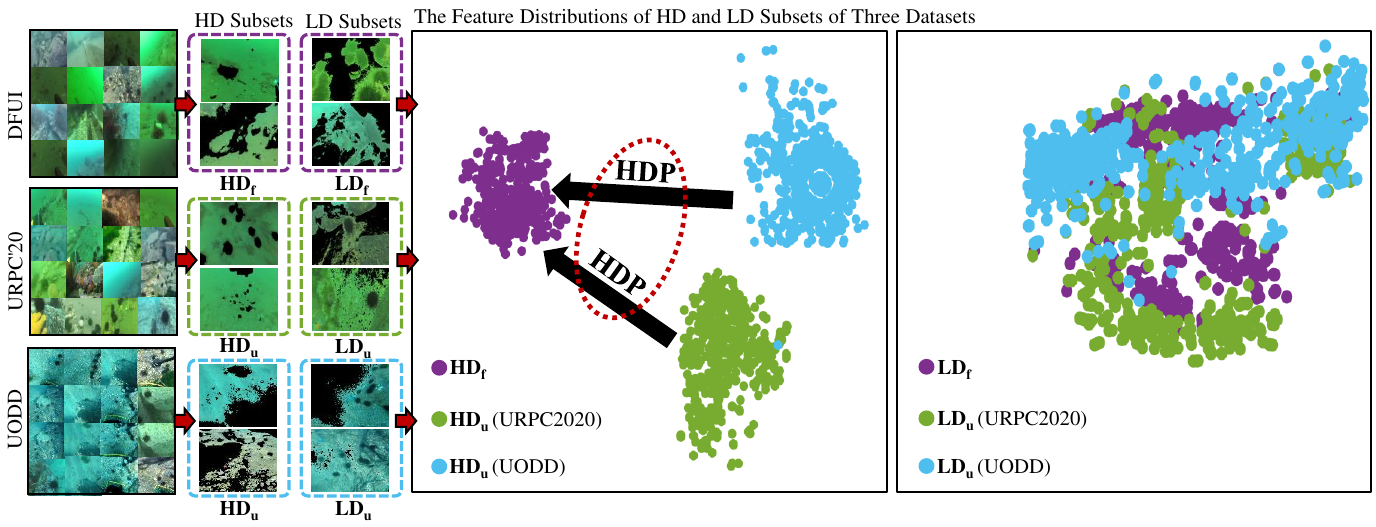}
	\caption{The overview of DFUI. Images of DFUI do not look sharp images to the eye. The gap between DFUI and underwater images lies mainly at the heavily instead of lightly degraded regions.}
	\label{DFUI}
\end{figure*}

\section{Related Work}
Visual object detection aims at determining what and where an object is in an image. Deep-learning based detectors generally consist of four cascaded parts: a backbone that extracts feature from an image, a neck that fuses multi-level features, an optional region proposal network that generates prediction candidates, and a head predicting classification and localization. There exist two major categories of solutions, ~\emph{i.e.}, enhancing based and re-training based, to the UOD task in literatures.  

\textbf{CNN- and Transformer- based Detectors.}
CNN-based detectors in mainstream evolution of detectors is promoted by several works. R-CNN \cite{R-CNN} is the first to show that CNN-based methods could lead to dramatic improvement in detection performance. Following R-CNN, Fast R-CNN \cite{fast}, Faster R-CNN \cite{fasterrcnn}, Cascade R-CNN \cite{Cascade}, and DetectoRS\cite{detetcors} achieve further performance improvement. Meanwhile, a series of proposal free detectors has been proposed to speed up the detection development. SSD \cite{SSD}, RetinaNet \cite{Focal}, RefineDet++ \cite{RefineDet++}, and FreeAnchor \cite{free} extract feature from inputs straightly for predicting through a unified stage. In the mean time, some arts concentrate on anchor-free direction, Grid RCNN \cite{grid}, YOLOX \cite{YOLOX}, YOLO \cite{yolo}, and FoveaBox \cite{FoveaBox} generates bounding boxes according to key- or center- points.

Over the past two years, there has been growing interest in using transformer for detection. Transformer-based have shown strong performance compared with CNN-based methods from the perspective of both accuracy and speed. DETR \cite{DETR} is a pioneer for transformer-based methods and redesigns the detection framework. DETR treats the detection task as an intuitive set prediction problem and eliminates traditional components such as anchor generation and non-maximum suppression. After DETR, some arts concentrate on improving specific parts such as transformer backbone \cite{SWIN, StS, PVTv1, PVTv2} and pre-training scheme \cite{YOLOS, UP-DETR}.

\textbf{UIE+UOD.} This type of method first uses UIE during preprocessing to improve the quality of underwater images (UI), and then to perform detection. UIE of VD-UIE+UOD chases high visual performance \cite{watergan,UEGAN,FUNIE} and leads to inconsistency with detection performance requirements. Therefore, evaluation of VD-UIE+UOD methods shows a significant drop in UOD accuracy. Among them, some VD-UIE+UOD  methods \cite{watergan,UEGAN} are trained on synthetic images due to lacking paired UI and its corresponding high-quality counterparts, posing weak adaptation to real-world UI. Prior works on underwater detection evidence that images favoring detectors are different from images favoring visibility \cite{chenlong,houminjun}. Therefore, UDD-UIE+UOD develops detection-driven features for UOD is gradually becoming an attractive research direction in the underwater vision community. While promising, existing  UDD-UIE+UOD methods require complex neural architecture and training process \cite{chenlong,colorUIE}.

\section{Proposed Method}
This section details our method, starting from our studies that the heavily-degraded regions of underwater- and DFUI- images have an evident distribution gap. Then, we elaborate a residual feature transference module to learn the mapping between the heavily-degraded regions of the detector-friendly and underwater image. Finally, we give a two-stage training scheme to optimize underwater HDP-oriented detection networks with RFTM.
\subsection{Our studies on Underwater Heavily-degraded Prior}\label{3.1}
The underwater detection task, which has no shortage of large image datasets, however, has not benefited from the full power of computer vision and deep learning methods, partly because imaging quality degradations mask many valuable features of a scene. Typical underwater detection methods strive for generating high-quality images or features by directly removing quality degradations. Unfortunately, these methods suffer poor performance because imaging factors are either unstable, too sensitive, or compounded \cite{Sea-Thru}. 
 
Unlike these approaches catering for high-quality images or features, our methods seek transferable prior knowledge from detector-friendly images. The prior knowledge guides detectors in removing degradations that interfere with detection. Specifically, from the existing underwater detection dataset, we construct a dataset named DFUI that favors detectors. DFUI is automatically built by a detector, avoiding the unstable step of removing environmental degradation and eliminating inconsistencies between detection and visual quality.

Now our focus converts to how to gain prior knowledge from such a large dataset. We elaborate a series of statistical experiments on the DFUI and underwater datasets and find an essential difference between the two datasets. We learn a mapping of the essential difference between the DFUI and underwater dataset. The mapping relationship is the prior knowledge we are looking for and can make DFUI as guidance for removing degradation interferences for detection. The whole exploration process is described as follows.        

\begin{figure*}[!t]
	\centering
	\includegraphics[width=6.5in]{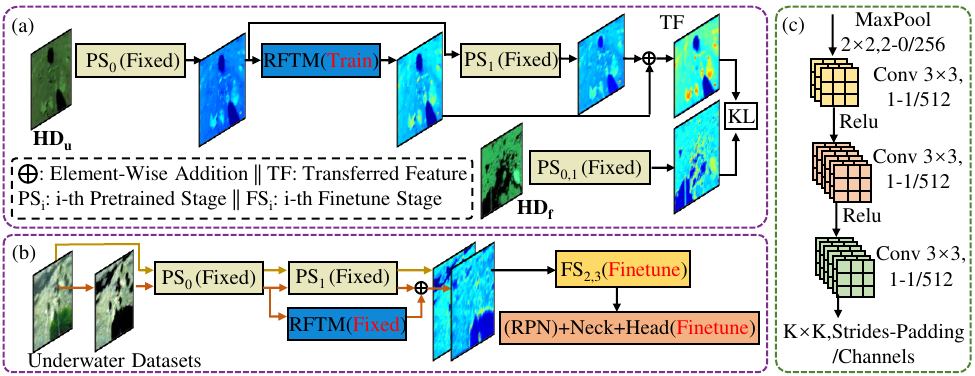}
	\caption{Two-stage training strategy. (a) the pipeline of the fist stage is training RFTM in an unsupervised manner on HD subset of underwater and DFUI datasets (\textit{i.e.,} $\mathbf{HD}_{\mathbf{u}}$ and $\mathbf{HD_{f}}$). (b) the pipeline of the second stage is finetuning the subsequent components of a detector on underwater datasets. (c) the structure of RFTM.}
	\label{f2}
\end{figure*}

For the training set of URPC2020 and URPC2021~\footnote{http://www.cnurpc.org/}, we first use Cascade RCNN to pick images with $\mathbf{AP\ge60}$ to construct DFUI datasets. Browsing the images of the constructed DFUI, we find an interesting phenomenon that detector-friendly images do not always have a positive correlation with visual quality. For example, in Fig.~\ref{DFUI}, there are many that do not look sharp images to the eye. It demonstrates that detectors do not always achieve more accurate detection on vision-friendly images. Therefore, it is necessary to build DFUI with detectors rather than artificial manner. 

Then, we constitute the HD and LD subsets of DFUI, URPC2020, and UODD according to the transmission value $\mathbf{t}$. In this paper, we use the traditional method UDCP \cite{UDCP} to estimate the transmission $\mathbf{t}$. This method \cite{UDCP} estimates transmission $\mathbf{t}$ depending on the underwater image and atmospheric light $\mathbf{A}$. The atmospheric light $\mathbf{A}$ is estimated by finding the brightest pixel in the underwater dark channel. Details can be referred to \cite{UDCP}.       



To obtain the feature representations of the HD and LD subsets in the embedding space \cite{RNGE}, we use ResNet50 pretrained on DFUI dataset as the feature extractor. The feature of each image was extracted from the second stage "Stage1", since the middle stage contains rich region information and the distribution of features is more pronounced. The t-SNE\cite{TSNE} technology is leveraged to visualize feature distribution. 

The feature distributions of the HD and LD subsets of DFUI, URPC2020, and UODD can be referred in Fig.~\ref{DFUI}. The results show that the gap between DFUI and underwater images lie mainly at the heavily instead of lightly degraded regions. Therefore, if we find a mapping between the heavily degraded regions of DFUI- and underwater- images, we can bride them together. At the heavily degraded regions, transferring the features from underwater to DFUI ones will certainly benefit the detection tasks. In this paper, we call such a mapping as the heavily degraded prior. Next, we will present an efficient solution to show how to learn and use the effective HDP.  


\subsection{Residual Feature Transference Module}
RFTM aims to learn the mapping relationship between the heavily degraded regions of DFUI- and underwater- images (\textit{i.e.,} $\mathbf{HD_{f}}$ and $\mathbf{HD}_{\mathbf{u}}$). It is inspired by the residual transfer network method proposed in \cite{resdual}. The gap between $\mathbf{HD_{f}}$ and $\mathbf{HD}_{\mathbf{u}}$ can be represented by a residual function that can be learned. We design a residual feature transference module (\textit{i.e.,} RFTM) to model the residual function. As is shown in Fig.~\ref{f2} (a), The RFTM is plugged into the $\mathbf{HD}_{\mathbf{u}}$ feed-forward pipeline. Through the $\mathbf{HD}_{\mathbf{u}}$ feed-forward pipeline with RFTM, we can get the transferred feature (TF) of $\mathbf{HD}_{\mathbf{u}}$. The TF is compared with its paired  $\mathbf{HD}_{\mathbf{f}}$ feature (i.e, the output of PS$_{0,1}$) to calculate the KL loss and the error is back-propagated to RFTM to update its parameters. We thus can explicitly learn the residual function and bridge the feature gap between $\mathbf{HD_{f}}$ and $\mathbf{HD}_{\mathbf{u}}$.

To include RFTM into the $\mathbf{HD}_{\mathbf{u}}$ feed-forward pipeline, we define the following feed-forward model. Let $\mathbf{\Delta} \mathbf{F}_\mathbf{s}$ be the proposed RFTM at the $\mathbf{s}$-th conv stage. For $\mathbf{\Delta} \mathbf{F}_\mathbf{s}$ the feed-forward equation at the $\mathbf{s}$-th conv stage can be formulated as, 

\begin{equation}\label{e0}
\begin{aligned}
\mathbf{\hat{F}}_\mathbf{s}(\mathbf{i})=\mathbf{F}_\mathbf{s}(\mathbf{i})+\mathbf{\Delta}\mathbf{F}_\mathbf{s}(\mathbf{F}_\mathbf{{s-1}}(\mathbf{i})),
\end{aligned}
\end{equation}
where $\mathbf{F}_\mathbf{s}(\mathbf{i})$ is the feature extracted from the $\mathbf{s}$-th stage for the input $\mathbf{i}$ sampled from $\mathbf{HD_{\mathbf{u}}}$ using the feature extraction network $\mathbf{F}$. $\mathbf{\Delta}\mathbf{F}_\mathbf{s}(\mathbf{F}_\mathbf{{s-1}}(\mathbf{i}))$ denotes the residual feature extracted from the output of $\mathbf{(s-1)}$-th stage, and $\mathbf{\hat{F}}_\mathbf{s}(\mathbf{i})$ denotes the feature extracted from the $\mathbf{s}$-th stage for any input $\mathbf{i} \in \mathbf{HD_{\mathbf{u}}}$ with RFTM modified feed-forward. 

The placement position of RFTM is illustrated in Fig.~\ref{f2}(a) the $\mathbf{HD_{\mathbf{u}}}$ feed-forward pipeline. It has no effect on $\mathbf{HD_{f}}$ feed-forward pipeline. Since the intermediate stage of the feature extraction network contains more regional information, the feature distribution gap is relatively large. In our case, we thus put RFTM at stage PS$_1$ of the feature extraction network to transfer the intermediate stage features between the heavily degraded regions of DFUI and underwater ones. 

How to design a suitable structure of RFTM plays a vital role in implementation. As mentioned above, RFTM is placed at stage PS$_1$ of the network and aims to transfer the intermediate stage features. Therefore, RFTM cannot over-extract the input features but ensures that the processed features contain as much region information as possible to conform to the characteristics of the intermediate stage feature. In addition, as underwater detectors are often deployed to mobile CPUs, it requires modules to be lightweight. For these purposes, we construct RFTM with few layers and small kernels. The structure of RFTM and its parameter settings are shown in Fig.~\ref{f2} (c). Specifically, RFTM contains a $\mathbf{2\times2}$ Maxpool layer with the stride of 2 and three $\mathbf{3\times3}$ convolutions with the stride of 1. "RELU" activation functions follow the first and second $\mathbf{3 \times 3}$ convolution, and there are no activation functions following the last $\mathbf{3 \times 3}$  convolution. 


\subsection{Two-stage training strategy}\label{3.3}
For efficiently plugging RFTM into a detector, we propose a two-stage learning scheme from the perspective of the unsupervised and finetune learning strategy. The two-stage training pipeline can be referred in Fig.~\ref{f2}(a)-(b). The first stage is training RFTM in an unsupervised manner on $\mathbf{HD_{f}}$ and $\mathbf{HD}_{\mathbf{u}}$ subsets without semantic labels. The second stage is fixed RFTM to finetune some components of a detector on underwater datasets.    

\textbf{Unsupervised transference training phase.} As is shown in Fig.~\ref{f2}(a), we use a pretrained network (on DFUI) to extract the low- and middle- level features of $\mathbf{HD}_{\mathbf{u}}$ and $\mathbf{HD_{f}}$. The low- and middle- level features are outputs of the shallow and intermediate stages of the network, respectively. For example, "Stage0" in ResNet50 \cite{ResNet}/ResNetXt101 \cite{ResNetXt} is the shallow stage, and "Stage1" in ResNet50 \cite{ResNet}/ResNetXt101 is the intermediate stage. The shallow and intermediate stages of the feature extraction network are fixed during the training phase and called PS$_0$, PS$_1$, and PS$_{0,1}$, respectively. The novel RFTM is used to accomplish the transference of the heavily degraded feature. In the $\mathbf{HD}_{\mathbf{u}}$ feed-forward pipeline, the output of PS$_0$ fed into RFTM. According to the residual feature transference strategy, the output feature from RFTM and PS$_1$ are element-wise added up. The resulting transferred feature (TF) is compared with the output of PS$_{0,1}$ to calculate the simple KL loss and the gradient is back-propagated to update RFTM. Let $\mathbf{j}$ be the image sampled from the HD subsets of DFUI (\textit{i.e.,} $\mathbf{HD_{\mathbf{f}}}$). The KL loss can be referred as follow,    
\begin{equation}
\label{eq:KL-loss}
\mathbf{L}_{\boldsymbol{kl}}(i,j)= \mathbf{F}_\mathbf{s}(\mathbf{j})\cdot \log \frac{\mathbf{F}_\mathbf{s}(\mathbf{j})} {\mathbf{\hat{F}}_\mathbf{s}(\mathbf{i})}.
\end{equation}
Here, the first stage is training the RFTM on the HD subsets $\mathbf{HD}_{\mathbf{u}}$ and $\mathbf{HD_{f}}$. After the first training stage, RFTM can transfer the features of heavily-degraded regions of underwater to detector-friendly ones. Note that we do not transfer the lightly-degraded regions of DFUI and underwater images due to no evident gap existing in lightly-degraded regions (details can be referred to Sec.~\ref{3.1}). In addition, $\mathbf{HD}_{\mathbf{u}}$ and $\mathbf{HD_{f}}$ are patches without semantic labels, and therefore the first training stage is an unsupervised process.        

\textbf{Finetune phase}. As is shown in Fig.~\ref{f2}(b), we put the trained RFTM into an existing feature extraction network, ~\emph{i.e.}, between its shallow and intermediate stage. Then, we finetune subsequent detection components, including high-level stages of the feature extraction network (denoted finetune stage (FS)), RPN, Neck, and Head, using common detection losses. The detection loss can be referred as follow,    
\begin{equation}
\centering
\label{eq:detection-loss}
\mathbf{L}_{\boldsymbol{det}}=\mathbf{L}_{\boldsymbol{bbox}}+\mathbf{L}_{\boldsymbol{cls}},
\end{equation}
where $\mathbf{L}_{\boldsymbol{bbox}}$ denotes the bounding-box regression loss and $\mathbf{L}_{\boldsymbol{cls}}$ denotes the object classification loss. The details of these individual loss components can be found in \cite{fasterrcnn}. Here, the second stage is finetune FS, RPN, Neck, and Head of detectors on training sets of underwater datasets. After the second stage, our methods can be tested on testing sets of underwater datasets.  

\section{Experiments}
In this section, the experiments described below aim to demonstrate the improvement of the perception performance of our proposed underwater heavily degraded prior for UOD, especially environmental degradations including haze-like effects, low visibility, and color distortions.

\subsection{Implementation Details}
Based on different feature extraction network, we build our detectors within Cascade R-CNN, denoted as RFTM-50 and RFTM-XT101 , respectively. Specifically, RFTM-50 plugs RFTM into CNN-based feature extraction network ResNet50; RFTM-XT101 plugs RFTM into CNN-based feature extraction network ResNetXT101 \footnote{ResNet50 and ResNetXT101 are popular feature extraction network, and most detectors support the two networks.}.

We train and test all the methods mentioned below using URPC2020 and UODD\cite{UODD} datasets. URPC2020 is a popular underwater dataset supporting by the China underwater robot professional contest, which contains 6,575 images in 4 categories (\textit{i.e.,} holothurian, echinus, scallop and starfish). We divide URPC2020 dataset into training and testing sets with a ratio of 7:3. In a word, there are totally 4,602 images for training and 1,973 images for testing. In order to verify that RFTM also has an effective performance improvement on other dataset, we carry out a comparative experiment on UODD. UODD is a rigorous benchmark with 2,688 images in 3 categories (\textit{i.e.,} holothurian, echinus and scallop) and widely used for underwater detection evaluation. UODD contains 2,560 images for training and 505 images for testing. 

\begin{table}[!t]
	\centering
	\caption{Accuracy comparison of RFTM with baselines in URPC2020 dataset. ($\mathbf{T=0.5}$).}
	\renewcommand{\arraystretch}{1.1}
	\label{URPC2020}
	\setlength{\tabcolsep}{0.7mm}{
		\begin{tabular}{llcccccc}
			\toprule 
			Methods&Backbone&AP&AP$_{50}$&AP$_{75}$&AP$_{S}$&AP$_{M}$&AP$_{L}$\\
			\midrule
			\textbf{Baseline:}&&&&&&&\\
			Cascade RCNN &ResNet50&45.0&78.8&47.5&18.7&39.3&50.7\\
			&ResNetXT101&45.2&80.3&46.4&22.7&41.6&50.7\\
			\midrule
			\textbf{UIE+UOD:}&&&&&&&\\
			FunIE+Grid RCNN&ResNetXT101&39.1&73.7&37.5&11.7&28.8&47.8\\
			FunIE+YOLOX&YOLOX-l&43.4&79.3&43.1&16.1&38.0&48.7\\
			FunIE+CSAM&DarkNet-53&43.8&76.0&40.8&19.3&44.4&48.1\\
			FunIE+Swin&Swin-B&46.3&77.2&47.8&21.2&43.3&50.2\\
			ERH+Swin&Swin-B&47.0&79.5&47.7&22.6&42.9&52.4\\
			\midrule
			\textbf{CNN:}&&&&&\\
			Free-anchor&ResNetXT101&44.4&80.0&44.6&24.3&41.7&49.3\\
			FoveaBox &ResNet101&43.2&79.3&42.8&24.2&40.2&48.1\\
			YOLOX &YOLOX-l&46.7&81.2&49.4&18.8&41.8&51.2\\
			Grid RCNN &ResNetXT101&43.5&78.4&44.1&23.6&40.3&48.9\\
			CSAM&DarkNet-53&46.4&79.2&41.1&20.4&44.7&50.8\\
			AquaNet&MNet&43.1&78.4&40.3&26.2&41.2&47.5\\
			\midrule
			\textbf{Transformer:}&&&&&\\
			PVTv1 &PVT-Medium&44.5&80.1&44.7&21.6&39.3&49.2\\
			PVTv2&PVTv2-B4&48.0&83.2&51.8&24.2&43.2&53.8\\
			Swin&Swin-B&50.2&83.0&54.4&24.9&44.7&55.6\\
			\midrule 
			\textbf{Ours:}&&&&&&&\\ 
			RFTM-50&ResNet50&48.2&80.7&50.0&19.5&41.6&53.1\\ 
			RFTM-XT101&ResNetXT101&{\color{red}{\textbf{50.9}}}&\color{red}\textbf{84.7}&{\color{red}{\textbf{55.2}}}&\color{red}\textbf{25.5}&{\color{red}{\textbf{45.1}}}&{\color{red}{\textbf{56.9}}}\\	
			\bottomrule
	\end{tabular}}
\end{table}

\begin{table}[!t]
	\centering
	\caption{Accuracy comparison of RFTM with baselines in UODD dataset. ($\mathbf{T=0.5}$).}
	\renewcommand{\arraystretch}{1.1}
	\label{UODD}
	\setlength{\tabcolsep}{0.7mm}{
		\begin{tabular}{llcccccc}
			\toprule
			Methods&Backbone&AP&AP$_{50}$&AP$_{75}$&AP$_{S}$&AP$_{M}$&AP$_{L}$\\
			\midrule
			\textbf{Baseline:}&&&&&&&\\
			Cascade RCNN&ResNet50&49.1&87.9&52.1&36.6&51.0&60.6\\
			&ResNetXT101&47.9&85.8&52.1&45.3&46.9&62.9\\
			\midrule
			\textbf{UIE+UOD:}&&&&&&&\\
			FunIE+Grid RCNN&ResNetXT101&36.2&73.1&31.7&18.4&35.5&55.6\\
			FunIE+YOLOX&YOLOX-l&47.1&83.8&47.8&30.4&46.0&62.5\\
			FunIE+CSAM&DarkNet-53&45.3&80.4&47.8&32.2&45.6&54.3\\
			FunIE+Swin&Swin-B&48.3&85.1&49.7&32.2&45.2&58.2\\
			ERH+Swin&Swin-B&48.8&86.0&48.5&30.6&48.6&61.3\\
			\midrule
			\textbf{CNN:}&&&&&&&\\
			Free-anchor&ResNetXT101&49.7&88.0&52.9&47.2&49.3&58.9\\
			FoveaBox&ResNet101&45.6&85.1&43.5&32.4&45.5&57.2\\
			YOLOX&YOLOX-l&48.8&86.3&51.7&36.7&47.6&63.0\\
			Grid RCNN&ResNetXT101&49.8&86.6&52.3&41.0&51.0&59.3\\
			CSAM&DarkNet-53&49.1&88.4&48.3&34.0&49.9&61.2\\
			AquaNet&MNet&45.2&84.3&44.0&30.5&44.2&55.2\\
			\midrule
			\textbf{Transformer:}&&&&&&&\\
			PVTv1 &PVT-Medium&45.8&85.4&43.6&32.0&45.8&56.9\\
			PVTv2 &PVTv2-B4&47.5&88.1&46.6&31.6&48.1&54.5\\
			Swin&Swin-B&51.0&89.4&53.5&36.5&50.5&61.8\\
			\midrule
			\textbf{Ours:}&&&&&&&\\
			RFTM-50&ResNet50&50.8&89.0&{\color{red}{\textbf{53.6}}}&33.6&50.9&62.8\\
			
			RFTM-XT101&ResNetXT101&{\color{red}{\textbf{52.7}}}&{\color{red}{\textbf{90.8}}}&50.0&{\color{red}{\textbf{47.7}}}&{\color{red}{\textbf{52.4}}}&{\color{red}{\textbf{63.5}}}\\	
			\bottomrule
	\end{tabular}}
\end{table}

All the detectors are trained on URPC2020/UODD training sets with pretrained models from DFUI dataset. Due to the different image sizes in the datasets, we resize all image sizes to $1333\times800$ for URPC2020 and $640\times640$ for UODD as input. We use a batch size of 2 and set the learning rate as 0.002 training on one NVIDIA A40. All the experiments are performed on MMDetection \footnote{https://github.com/open-mmlab/mmdetection}.

For testing process, we keep boxes with confidence threshold greater than 0.3 for subsequent evaluation. For the accuracy evaluation, we adopt the standard COCO metrics (mean average precision, \textit{i.e.,} mAP). AP, AP$_{50}$, and AP$_{75}$ mean the mAP at IoU=[0.50:0.05:0.95], 0.50, and 0.75, respectively. AP$_{S}$, AP$_{M}$, and AP$_{L}$ mean the AP for objects of area smaller than $32\times32$, between $32\times32$ and $96\times96$, larger than $96\times96$, respectively. For the efficiency evaluation, we adopt two metrics, \textit{i.e.,} Param. and FPS. Param. is the parameters of a detector and FPS is frames per second.

\subsection{Comparison with Advanced Detectors}

\quad \textbf{Comparison with the baseline.} We compare our methods with the baseline detectors Cascade RCNN. On URPC2020, as can be seen from Table~\ref{URPC2020}, our method reveals a significant improvement and wins on all accuracy metrics. For instance, compared with Cascade R-CNN (ResNet50), RFTM-50 outperforms it by 3.2\% in AP, 2.5\% in AP$_{75}$, and 2.3\% in AP$_{M}$. Compared with Cascade R-CNN (ResNetXT101), RFTM-XT101 outperforms it by 5.7\% in AP, 4.4\% in AP$_{50}$, and 8.8\% in AP$_{75}$. On UODD, Table~\ref{UODD} show that our methods also has similar marked improvements. For instance, compared with Cascade R-CNN (ResNet50), RFTM-50 outperforms it by 1.7\% in AP, 1.1\% in AP$_{50}$, and 1.5\% in AP$_{75}$. Compared with Cascade R-CNN (ResNetXT101), RFTM-XT101 outperforms it by 4.8\% in AP, 5.0\% in AP$_{50}$, and 5.5\% in AP$_{M}$. These results demonstrate the superiority of our method.   

\textbf{Comparison with advanced detectors.} On URPC2020 and UODD, we conducted several experiments on other advanced detectors including UIE+UOD and CNN-/Transformer- besed methods. For UIE+UOD detectors, we select common underwater enhancement methods FunIE-GAN \cite{FUNIE} and ERH \cite{ERH} as the pre-processing methods. URPC2020 and UODD are pre-processed by FunIE-GAN/ERH and then send to detectors (Grid R-CNN, YOLOX, CSAM, and Swin) for training and testing. For the CNN-based methods, we select Free-anchor \cite{free}, FoveaBox \cite{FoveaBox}, YOLOX \cite{YOLOX}, Grid RCNN \cite{grid}, CSAM \cite{UODD}, and AquaNet \cite{liuhongwei} as comparisons. For the Transformer-based methods, we select Pvtv1 \cite{PVTv1}, Pvtv2 \cite{PVTv2}, and Swin Transformer \cite{SWIN} as comparisons. As can be seen from Table~\ref{URPC2020} and Table~\ref{UODD}, our methods significantly surpass these detectors.

On URPC2020, UIE+UOD uses underwater enhancement methods to enhance images for subsequent detection. The performance of these methods on detection tasks is disappointing. \cite{chenlong} reports that UIE+UOD has inconsistent pursuits, the former pursuing image quality while the latter pursuing detection accuracy, leading to disappointing detection results.

CNN-based detectors are mainstream methods in the detection community, and most existing UOD methods are developed based on CNN ones. Our methods outperform CNN-based detectors by a large margin. Free-anchor and Grid RCNN adopt ResNetXT101 as the feature extraction network. Our RFTM-XT101 outperforms Free-anchor and Grid R-CNN by 6.5\% and 7.4\% in AP, respectively. RFTM-XT101 also outperforms YOLOX by 4.2\% in AP. RFTM-50 achieves 48.2\% in AP and is 5.0\% higher than FoveBox. \textit{CSAM fuses high-level image information for UOD tasks while neglects removing underwater degradations.} RFTM-50 surpasses CSAM by 1.8\% in AP, 1.5\% in AP$_{50}$, and 8.9\% in AP$_{75}$. RFTM-XT101 surpasses CSAM by 4.5\% in AP, 5.5\% in AP$_{50}$, and 14.1\% in AP$_{75}$. \textit{AquaNet designs a Multi-scale Contextual block and Multi-scale Blursampling block for their backbones to improve the performance of underwater detection.} RFTM-50 surpasses AquaNet by 5.1\% in AP, 2.3\% in AP50, and 9.7\% in AP75. Similarly, RFTM-XT101 also achieves better performance than AquaNet by a large margin.

Transformer based detector has risen in recent years and attracted much focus due to its excellent detection performance. Our methods still surpass these types of detectors. Typically, RFTM-XT101 achieves 50.9\% in AP and is 0.7\% higher than Swin which is best in the baseline. For AP$_{50}$, RFTM-XT101 reaches 84.7\% and outperforms Swin by 1.7\%. For AP$_{75}$, RFTM-XT101 reaches 55.2\% and outperforms Swin by 0.8\%. 

\begin{table}[!t]
	\centering
	\caption{Accuracy performance of various detectors on DFUI dataset. DFUI benefits various detectors.}
	\renewcommand{\arraystretch}{1.4}
	\label{DFUI-preformance}
	\setlength{\tabcolsep}{5mm}{
		\begin{tabular}{lcc}
			\Xhline{0.8pt}  
			Methods&AP&AP$_{50}$\\
			\Xhline{0.5pt}	
			Cascade R-CNN&79.7&98.0\\
			
			FoveaBox&76.6&98.3\\
			
			YOLOX&75.9&97.9\\
			
			Grid R-CNN&77.1&97.9\\
			
			DetectoRS&76.3&98.9\\
			
			Swin Transformer&72.4&95.3\\
			
			\Xhline{0.8pt} 
	\end{tabular}}
\end{table}

\begin{table}[!t]
	\centering
	\caption{On URPC2020 dataset, AP results of RFTM-50 and RFTM-XT101 with various threshold value $\mathbf{T}$.}
	\renewcommand{\arraystretch}{1.5}
	\label{T-URPC2020}
	\setlength{\tabcolsep}{0.6mm}{
		\begin{tabular}{lcccccccccc}
			\toprule 
			$\mathbf{T}$ value&0.1&0.2&0.3&0.4&0.5&0.6&0.7&0.8&0.9&1.0\\
			\midrule
			RFTM-50&48.7&48.5&47.8&49.1&48.2&49.1&{\color{red}{\textbf{49.5}}}&49.1&49.3&48.8\\
			RFTM-XT101&49.8&49.3&48.9&50.3&{\color{red}{\textbf{50.9}}}&50.4&50.0&50.0&50.2&50.1\\
			\bottomrule
	\end{tabular}}
\end{table}

\begin{table}[!t]
	\centering
	\caption{On UODD dataset, AP results of RFTM-50 and RFTM-XT101 with various threshold value $\mathbf{T}$.}
	\renewcommand{\arraystretch}{1.5}
	\label{T-UODD}
	\setlength{\tabcolsep}{0.6mm}{
		\begin{tabular}{lcccccccccc}
			\toprule 
			$\mathbf{T}$ value&0.1&0.2&0.3&0.4&0.5&0.6&0.7&0.8&0.9&1.0\\
			\midrule
			RFTM-50&49.8&48.4&50.6&50.9&50.8&{\color{red}{\textbf{51.3}}}&51.0&51.1&51.1&50.9\\
			RFTM-XT101&52.4&52.8&52.7&52.8&52.7&53.0&{\color{red}{\textbf{53.8}}}&53.1&53.4&50.4\\
			\bottomrule
	\end{tabular}}
\end{table} 

\begin{figure}[!t]
	\centering
	\includegraphics[width=3.5in,height=1.7in]{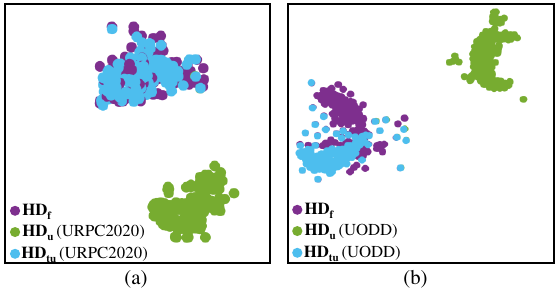}
	\caption{Transference performance results of RFTM. (A) is the feature distributions of HD sub sets on URPC2020. (B) is the feature distributions of HD sub sets on UODD. Here, we adopt RFTM-50, $\mathbf{T}=\mathbf{0.7}$ for URPC2020 and $\mathbf{0.6}$ for UODD. At the value $\mathbf{T}=\mathbf{0.7,0.6}$, RFTM-50 achieves the best improvement on URPc2020 and UODD, respectively.}
	\label{RFTM}
\end{figure} 

On UODD, our methods also achieve the best performance among all detectors and have significant improvements compared with others. These experimental results further demonstrate the superiority of our method.

\textbf{Comparison on accuracy-speed-size trade-off.} On URPC2020, we provide the comparison results in terms of accuracy-speed-size trade-off of accurate models. As is shown in Fig.~\ref{contri0}(c), compared with other detectors, our methods achieve the best compromise results. For instance, RFTM-50 surpasses Cascade R-CNN, Grid R-CNN, and PVTv2 with fewer parameters and higher speeds. For similar parameter sizes, RFTM-50 surpasses Free-anchor and PVTv1 by a large margin with higher speeds. RFTM-XT101 surpasses Swin with fewer parameters and higher speeds.

\begin{table}
	\centering
	\caption{Brief of various training strategies. UI: Underwater images, TF: Transference training, FT: Finetune, and *$_{r}$: Randomly generate patch subsets for *.}
	\renewcommand{\arraystretch}{1.4}
	\label{Ttraing strategy}
	\setlength{\tabcolsep}{1.3mm}{
		\begin{tabular}{l|cc|ccccc}
			\Xhline{0.8pt}  
			\multirow{2}{*}{Strategy}&\multicolumn{2}{c|}{Stage}&\multicolumn{5}{c}{Patch}\\
			
			&TF&FT&$\mathbf{HD}_\mathbf{u}$&$\mathbf{UI}_\mathbf{r}$&DFUI&$\mathbf{HD}_\mathbf{f}$&$\mathbf{DFUI}_\mathbf{r}$\\
			\Xhline{0.5pt} 
			
			CAS&-&-&-&-&-&-&-\\
			
			CAS+${\mathbf{HD}_\mathbf{u}}$&-&-&\ding{51}&-&-&-&- \\
			
			CAS+${\mathbf{UI}_\mathbf{r}}$&-&-&-&\ding{51}&-&-&- \\
			
			\Xhline{0.5pt} 
			
			CAS+TF1&\ding{51}&\ding{51}&-&\ding{51}&-&-&\ding{51} \\
			
			CAS+TF2&\ding{51}&\ding{51}&\ding{51}&-&\ding{51}&-&- \\
			
			no FT&\ding{51}&-&\ding{51}&-&-&\ding{51}&- \\
			\Xhline{0.5pt}
			RFTM-50&\ding{51}&\ding{51}&\ding{51}&-&-&\ding{51}&- \\
			\Xhline{0.8pt} 
	\end{tabular}}
\end{table}

\begin{table}
	\centering
	\caption{Accuracy results of various training strategies on URPC2020. The based feature extraction network is ResNet50. $\mathbf{T}=\mathbf{0.5}$.}
	\renewcommand{\arraystretch}{1.4}
	\label{Ttraing strategy-R}
	\setlength{\tabcolsep}{5mm}{
		\begin{tabular}{lccc}
			\Xhline{0.8pt}  
			Strategy&AP&AP$_{50}$&AP$_{75}$\\
			\Xhline{0.5pt}	
			CAS&45.0&78.8&47.5\\
			
			CAS+${\mathbf{HD}_\mathbf{u}}$&24.3&49.4&21.3\\
			
			CAS+${\mathbf{UI}_\mathbf{r}}$&21.7&43.4&25.2\\
			
			\Xhline{0.5pt}
			
			CAS+TF1&46.6&78.8&49.8\\
			
			CAS+TF2&47.2&80.2&45.5\\
			
			no FT&47.6&79.7&49.2\\
			
			\Xhline{0.5pt}
			RFTM-50&\color{red}\textbf{48.2}&\color{red}\textbf{80.7}&\color{red}\textbf{50.0}\\
			
			\Xhline{0.8pt} 
	\end{tabular}}
\end{table}

\subsection{Performance Analysis}\label{4.3}
\textbf{Generalization of DFUI.} To evaluate the generalization of DFUI, we observe the accuracy performance of various detectors on DFUI dataset. We select 6 popular detectors from CNN-/Transformer- based methods, including Cascade R-CNN, FoveaBox, YOLOX, Grid R-CNN, DetectoRS, and Swin Transformer. as shown in Table~\ref{DFUI-preformance}, all methods achieve high AP and AP$_{50}$ and surpass their performance on common underwater datasets by a large margin. These results demonstrate that DFUI set is beneficial to various detectors although only built by one detector. 

\textbf{Influence on different Threshold $\mathbf{T}$.} We first explore the influence of the medium transmission $\mathbf{t}$ on underwater heavily degraded prior modeling. We use different threshold T to generate various $\mathbf{HD}_{\mathbf{u}}$ and $\mathbf{HD_{f}}$. In this paper, we set $\mathbf{T}$ as $\mathbf{(0.1, 0.2, 0.3, \cdots, 1.0)}$ and conduct explorations on URPC2020 and UODD. 

\begin{figure*}[!t]
	\centering
	\includegraphics[width=7.2in,height=3.2in]{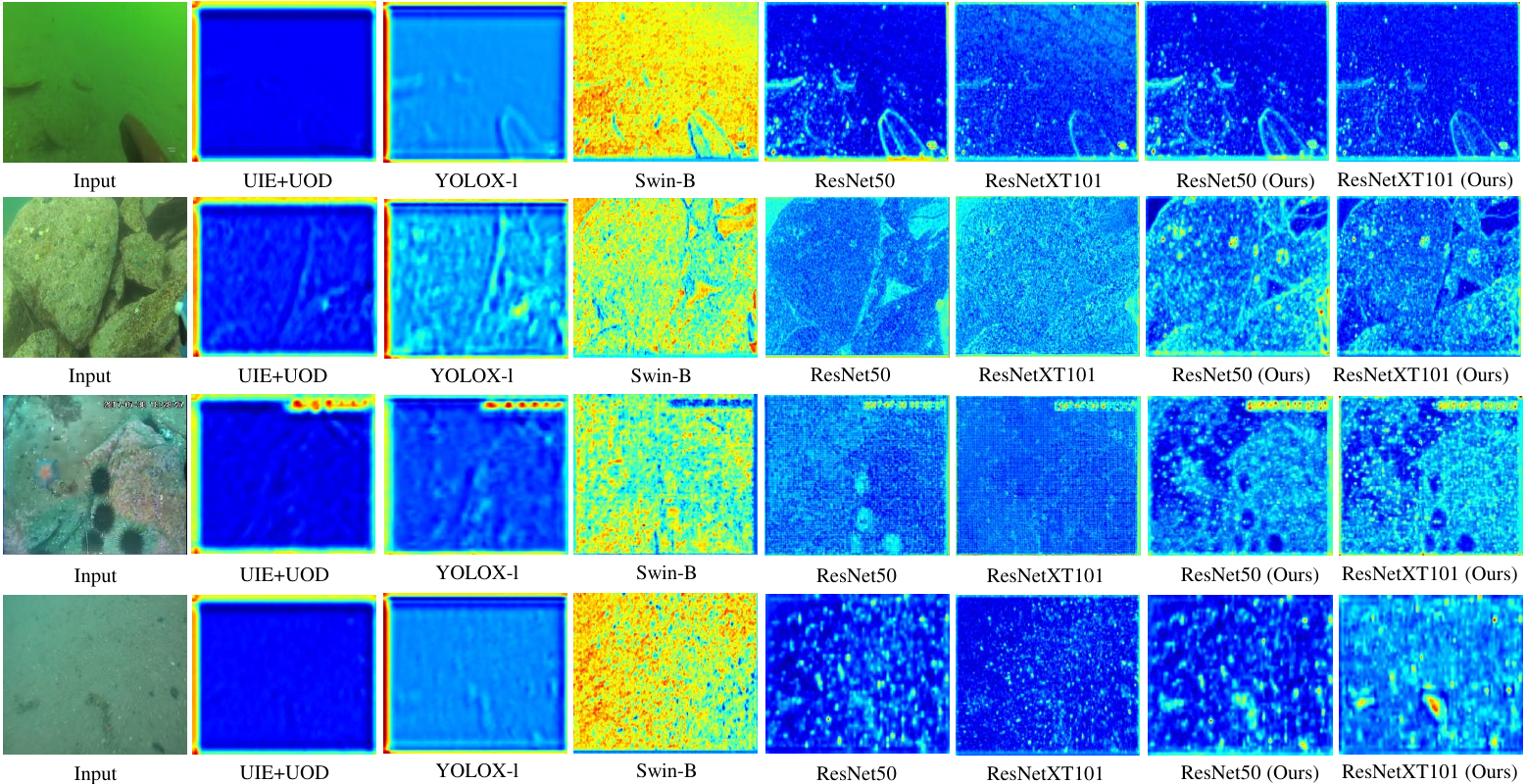}
	\caption{Examples of visualization of the feature maps on URPC2020. The feature maps are extracted from the Stage1 of these feature extraction networks. The example from top to bottom is haze-like effects, color distortions, and low visibility, respectively.}
	\label{f3}
\end{figure*}

\begin{table}[!t]
	\centering
	\caption{Brief of various parameter settings of RFTM.}
	\renewcommand{\arraystretch}{1.5}
	\label{RFTM-sturcture}
	\setlength{\tabcolsep}{5mm}{
		\begin{tabular}{cccc}
			\toprule
			Settings&Kernel&Layers&Position\\
			\midrule
			Ker5&$5\times5$&-&-\\
			Layr4&-&4&-\\
			Pos0&-&-&Before Stage$_0$\\
			Pos1&-&-&After Stage$_1$\\
			Pos2&-&-&After Stage$_2$\\
			\midrule
			RFTM-50&$3\times3$&3&After Stage$_0$\\
			\bottomrule
			
	\end{tabular}}
\end{table} 

\begin{table}[!t]
	\centering
	\caption{Experiment results in various parameter settings of RFTM on URPC2020. The based feature extraction network is ResNet50. $\mathbf{T=\mathbf{0.5}}$.}
	\renewcommand{\arraystretch}{1.5}
	\label{RFTM-result}
	\setlength{\tabcolsep}{1.5mm}{
		\begin{tabular}{ccccccc}
			\toprule
			Settings&Ker5&Layer4&Pos0&Pos1&Pos2&RFTM-50\\
			\midrule
			AP&45.5&45.2&46.8&46.9&46.2&\color{red}\textbf{48.2}\\
			
			\bottomrule
			
	\end{tabular}}
\end{table}

The results are reported in Table~\ref{T-URPC2020} and Table~\ref{T-UODD}, respectively. As can be seen, the performance of RFTM is somewhat sensitive to the degrees of degradation. For instance, on URPC2020, RFTM-50 with $\mathbf{T=0.7}$ achieves 49.5\% in AP and is 1.7\% higher than $\mathbf{T=0.3}$. RFTM-XT101 with $\mathbf{T=0.5}$ achieves 50.9\% in AP and is 2.0\% higher than $\mathbf{T=0.3}$. On UODD, RFTM-50 with $\mathbf{T=0.6}$ achieves 51.3\% in AP and is 2.9\% higher than $\mathbf{T=0.2}$. RFTM-XT101 with $\mathbf{T=0.7}$ achieves 53.8\% in AP and is 3.4\% higher than $\mathbf{T=1.0}$. In addition, the best results tend to be achieved in the medium $\mathbf{T}$ value, \textit{i.e.,} $\mathbf{T=0.5, 0.6,}$ or $\mathbf{0.7}$. While the extreme degradation condition that $\mathbf{T=0.2/0.3/1}$, the improvement is not noticeable. We deduce that the smaller the value $\mathbf{T}$, the more regions that denoted as lightly-degraded. This results in regions that should be transferred being divided into regions that do not need transference, leading to poor improvements. While the higher the value $\mathbf{T}$, the more regions that denoted as highly-degraded. This results in regions that should not be transfered being divided into regions that need to be transfered leading to poor improvements. Therefore, we recommend choosing the medium Threshold $\mathbf{T}$. 

\textbf{Transference performance of RFTM.} RFTM aims to transferring highly-degraded regions of DFUI and underwater ones. To evaluate the transference performance of the RFTM, we visualizes the feature distribution of three HD subsets, \textit{i.e.,} $\mathbf{HD_{f}}$, $\mathbf{HD}_{\mathbf{u}}$, and $\mathbf{HD}_{\mathbf{tu}}$. $\mathbf{HD}_{\mathbf{tu}}$ denotes $\mathbf{HD}_{\mathbf{u}}$ processed by RFTM. As is shown in Fig.~\ref{RFTM}, there exist overlaps between $\mathbf{HD_{f}}$ and $\mathbf{HD}_{\mathbf{tu}}$, while there remain exist evident gaps between $\mathbf{HD_{f}}$ and $\mathbf{HD}_{\mathbf{u}}$. The feature distribution demonstrates that RFTM complete the transference task on $\mathbf{HD}_{\mathbf{u}}$ to  $\mathbf{HD_{f}}$.

\begin{figure*}[!t]
	\centering
	\includegraphics[width=7in,height=3.2in]{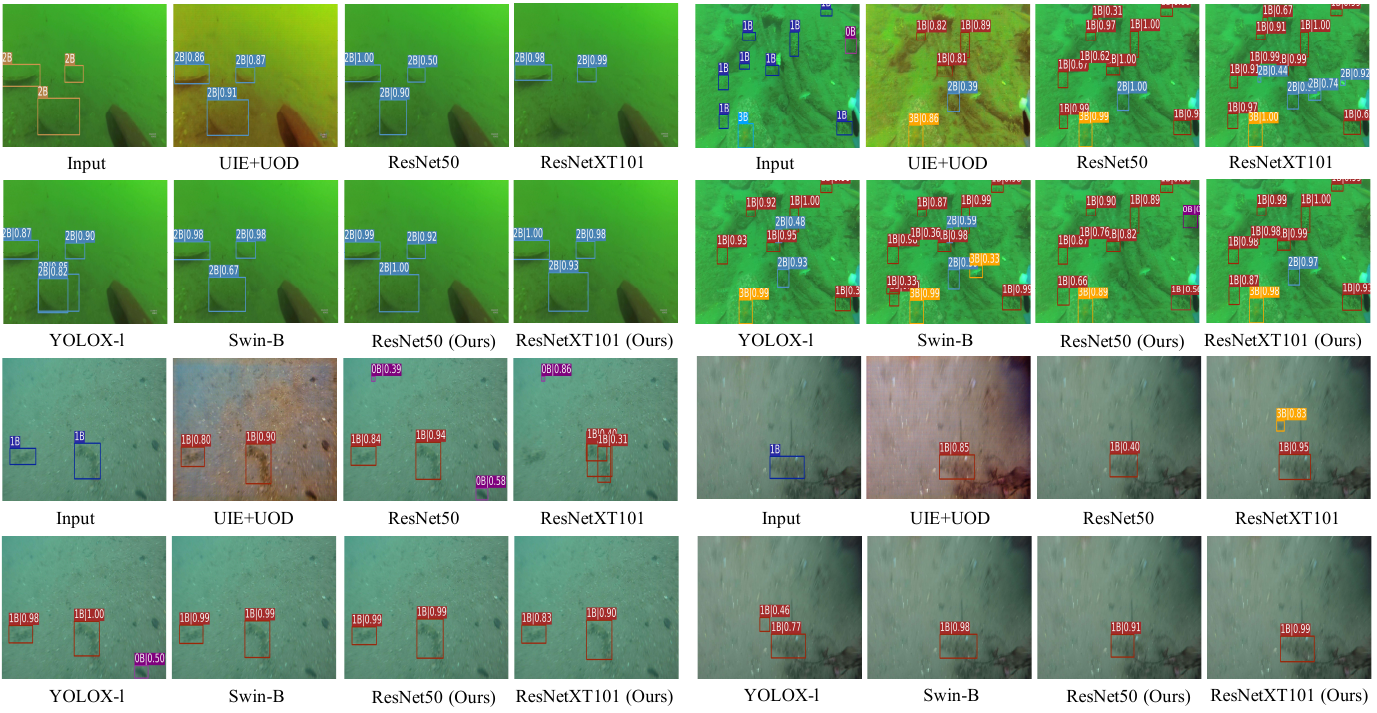}
	\caption{Some qualitative examples on URPC2020. The example from left to right, top to bottom, is haze-like effects, color distortions, and low visibility, respectively.}
	\label{f4}
\end{figure*}

\textbf{Influence on training strategy.} We perform an ablation study where we evaluate the following configurations to analyze the effectiveness of different training strategies in our methods. As is shown in Table~\ref{Ttraing strategy}, we progressively add additional steps which enables us to gauge the performance improvements obtained by each of them,
\begin{itemize}
	\item CAS: the baseline experiment where Cascade R-CNN is trained on the training set of URPC2020.  
	\item CAS+${\mathbf{HD}_\mathbf{u}}$: Cascade R-CNN is trained on the training set and HD subset of URPC2020. 
	\item CAS+${\mathbf{UI}_\mathbf{r}}$: Cascade R-CNN is trained on the training set and $\mathbf{UI}_\mathbf{r}$ subset of URPC2020. $\mathbf{UI}_\mathbf{r}$ is constituted by randomly masked URPC2020 images. 
	\item CAS+TF1: RFTM-50 train RFTM on the $\mathbf{UI}_\mathbf{r}$ subset of URPC2020 and $\mathbf{DFUI}_\mathbf{r}$ subset of DFUI. $\mathbf{DFUI}_\mathbf{r}$  is constituted by randomly masked DFUI images.  
	\item CAS+TF2: RFTM-50 train RFTM on the $\mathbf{HD}_\mathbf{u}$ and DFUI.
	\item no FT: RFTM-50 do not finetune the subsequent components of detectors.
	\item RFTM-50: our ultimate training program which can be referred to Sec.~\ref{3.3}. 
\end{itemize}

As can be seen in Table~\ref{Ttraing strategy-R}, CAS+${\mathbf{HD}_\mathbf{u}}$ and CAS+${\mathbf{UI}_\mathbf{r}}$ use patches and underwater images (here, URPC2020) train Cascade R-CNN, however, their performance on detection task is disappointing. Compared with CAS, CAS+TF1, CAS+TF2, and no FT use DFUI related datasets training the proposed RFTM, they have improved to varying degrees. However, RFTM-50 outperforms CAS+TF1 by 1.6\%, CAS+TF2 by 1\%, and no FT by 0.6\% in AP. These results indicates that it is important to transfer features under the guidance of transmission $\mathbf{t}$. More importantly, finetune can further improve the perception performance.

\textbf{Structure of RFTM.}  We perform an ablation study where we evaluate the following configurations to analyze the effectiveness of different parameter settings in RFTM. As is shown in Table~\ref{RFTM-sturcture}, we progressively change the setting which enables us to gauge the performance obtained by each of them,
\begin{itemize}
	\item Ker5: adopting $5\times5$ kernels.  
	\item Layer4: adopting five $3\times3$ convolutions.
	\item Pos0: putting RFTM before Stage$_0$ of ResNet50. 
	\item Pos1: putting RFTM after Stage$_1$ of ResNet50. 
	\item Pos2: putting RFTM after Stage$_2$ of ResNet50.
\end{itemize}

As can be seen from Table~\ref{RFTM-result}, RFTM-50 outperforms Ker5 by 2.7\% and Layer4 by 3\% in AP, respectively. It demonstrates that the property of RFTM does not benefit from large receptive fields. RFTM-50 is higher than Pos0, Pos1 and Pos2. It demonstrates that transferring the intermediate stage feature is the most efficient setting.

\textbf{Environmental degradation performance.} We present the feature visualization results to demonstrate the performance of our methods on various environmental degradations. Here we take images from URPC2020 as examples. We visualize the feature maps of UIE+UOD (FUnIE+YOLOX), ResNet50 (Cascade R-CNN), ResNetXT101 (Cascade R-CNN), YOLOX-l (YOLOX), Swin-B (Swin Transformer), and ours. For haze-like effects, color distortions, and low visibility, the feature response of objects of UIE+UOD and YOLOX-l is relatively weak. The amplitude of the feature response of objects of ResNet50, ResNetXT101, and YOLOX-l is attenuated inconsistently. But our methods significantly boost the feature response on the discriminative region while suppressing the interference.   

We also present some qualitative results to further demonstrate the performance of our methods on various environmental degradations. For color distortions, most methods fail to complete detection, there are errors and missed detection phenomena of these feature extraction networks. In contrast, our Cascade RFTM-50 completes the detection correctly. For haze-like effects and low visibility, most and our methods can complete the detection task very well. Only a few methods have error and missed detection phenomena, for example, YOLOX-l has error detection. ResNet50 and ResNetXT101 have error and missed detection phenomena. The qualitative results also demonstrate that our method actually does perform well on various environmental degradations.

\section{Conclusion}
In this paper, we introduce a novel heavily degraded prior (HDP) for low-quality UOD task, which can be used to reduce the feature mismatch between the heavily degraded regions of underwater- and DFUI- images. We design a simple and effective module named residual feature transference module (RFTM) to learn the HDP by adaptively transferring features at heavily degraded regions. The proposed RFTM can be easily learned without the supervision of semantic labels and plugged into existing popular CNN-based feature extraction networks to improve their performance. In addition, we propose a novel two-stage training strategy to coordinating the feature extraction networks with RFTM. Extensive comparison and analysis experiments conducted on two popular UOD datasets, \textit{i.e.,} URPC2020 and UODD, our methods significantly boost the detection performance (especially various environmental degradations) without bells and whistles.

\section{ACKNOWLEDGEMENT}    
This work was supported in part by the National Key R\&D Program of China under Grant 2020YFB1313503; in part by the National Natural Science Foundation of China under Grants 61922019, 61733002, 62027826 and 61772105; in part by the LiaoNing Revitalization Talents Program under Grant XLYC1807088; and in part by the Fundamental Research Funds for the Central Universities.

\bibliographystyle{IEEEtran}
\bibliography{egbib}

\begin{thebibliography}{10}
\providecommand{\url}[1]{#1}
\csname url@samestyle\endcsname
\providecommand{\newblock}{\relax}
\providecommand{\bibinfo}[2]{#2}
\providecommand{\BIBentrySTDinterwordspacing}{\spaceskip=0pt\relax}
\providecommand{\BIBentryALTinterwordstretchfactor}{4}
\providecommand{\BIBentryALTinterwordspacing}{\spaceskip=\fontdimen2\font plus
\BIBentryALTinterwordstretchfactor\fontdimen3\font minus
  \fontdimen4\font\relax}
\providecommand{\BIBforeignlanguage}[2]{{%
\expandafter\ifx\csname l@#1\endcsname\relax
\typeout{** WARNING: IEEEtran.bst: No hyphenation pattern has been}%
\typeout{** loaded for the language `#1'. Using the pattern for}%
\typeout{** the default language instead.}%
\else
\language=\csname l@#1\endcsname
\fi
#2}}
\providecommand{\BIBdecl}{\relax}
\BIBdecl

\bibitem{houminjun}
R.~Liu, X.~Fan, M.~Zhu, M.~Hou, and Z.~Luo, ``Real-world underwater
  enhancement: Challenges, benchmarks, and solutions under natural light,''
  \emph{IEEE TCSVT}, vol.~30, no.~12, pp. 4861--4875, 2020.

\bibitem{chenlong}
L.~Chen, Z.~Jiang, L.~Tong, Z.~Liu, A.~Zhao, Q.~Zhang, J.~Dong, and H.~Zhou,
  ``Perceptual underwater image enhancement with deep learning and physical
  priors,'' \emph{IEEE TCSVT}, vol.~31, no.~8, pp. 3078--3092, 2021.

\bibitem{PDCNet}
X.~Chen, H.~Li, Q.~Wu, K.~N. Ngan, and L.~Xu, ``High-quality {R-CNN} object
  detection using multi-path detection calibration network,'' \emph{IEEE
  TCSVT}, vol.~31, no.~2, pp. 715--727, 2021.

\bibitem{DRnet}
X.~Chen, J.~Yu, S.~Kong, Z.~Wu, and L.~Wen, ``Joint anchor-feature refinement
  for real-time accurate object detection in images and videos,'' \emph{IEEE
  TCSVT}, vol.~31, no.~2, pp. 594--607, 2021.

\bibitem{ESCNet}
J.~Nie, Y.~Pang, S.~Zhao, J.~Han, and X.~Li, ``Efficient selective context
  network for accurate object detection,'' \emph{IEEE TCSVT}, vol.~31, no.~9,
  pp. 3456--3468, 2021.

\bibitem{RefineDet++}
S.~Zhang, L.~Wen, Z.~Lei, and S.~Z. Li, ``Refinedet++: Single-shot refinement
  neural network for object detection,'' \emph{IEEE TCSVT}, vol.~31, no.~2, pp.
  674--687, 2021.

\bibitem{FUNIE}
M.~J. Islam, Y.~Xia, and J.~Sattar, ``Fast underwater image enhancement for
  improved visual perception,'' \emph{{IEEE} Robotics Autom. Lett.}, vol.~5,
  no.~2, pp. 3227--3234, 2020.

\bibitem{jiang2022target}
Z.~Jiang, Z.~Li, S.~Yang, X.~Fan, and R.~Liu, ``Target oriented perceptual
  adversarial fusion network for underwater image enhancement,'' \emph{IEEE
  Transactions on Circuits and Systems for Video Technology}, 2022.

\bibitem{liu2022twin}
R.~Liu, Z.~Jiang, S.~Yang, and X.~Fan, ``Twin adversarial contrastive learning
  for underwater image enhancement and beyond,'' \emph{IEEE Transactions on
  Image Processing}, vol.~31, pp. 4922--4936, 2022.

\bibitem{MU22}
P.~Mu, H.~Qian, and C.~Bai, ``Structure-inferred bi-level model for underwater
  image enhancement,'' in \emph{ACM MM}, 2022, pp. 2286--2295.

\bibitem{colorUIE}
C.-H. Yeh, C.-H. Lin, L.-W. Kang, C.-H. Huang, M.-H. Lin, C.-Y. Chang, and
  C.-C. Wang, ``Lightweight deep neural network for joint learning of
  underwater object detection and color conversion,'' \emph{IEEE TNNLS}, pp.
  1--15, 2021.

\bibitem{Cascade}
Z.~Cai and N.~Vasconcelos, ``Cascade {R-CNN:} delving into high quality object
  detection,'' in \emph{CVPR}, 2018, pp. 6154--6162.

\bibitem{YOLOX}
Z.~Ge, S.~Liu, F.~Wang, Z.~Li, and J.~Sun, ``{YOLOX:} exceeding {YOLO} series
  in 2021,'' \emph{CoRR}, vol. abs/2107.08430, 2021.

\bibitem{YOLOS}
Y.~Fang, B.~Liao, X.~Wang, J.~Fang, J.~Qi, R.~Wu, J.~Niu, and W.~Liu, ``You
  only look at one sequence: Rethinking transformer in vision through object
  detection,'' in \emph{NIPS}, 2021.

\bibitem{SWIN}
Z.~Liu, Y.~Lin, Y.~Cao, H.~Hu, Y.~Wei, Z.~Zhang, S.~Lin, and B.~Guo, ``Swin
  transformer: Hierarchical vision transformer using shifted windows,'' in
  \emph{ICCV}, 2021, pp. 9992--10\,002.

\bibitem{detetcors}
S.~Qiao, L.~Chen, and A.~L. Yuille, ``Detectors: Detecting objects with
  recursive feature pyramid and switchable atrous convolution,'' in
  \emph{CVPR}, 2021, pp. 10\,213--10\,224.

\bibitem{FoveaBox}
T.~Kong, F.~Sun, H.~Liu, Y.~Jiang, L.~Li, and J.~Shi, ``Foveabox: Beyound
  anchor-based object detection,'' \emph{IEEE TIP}, vol.~29, pp. 7389--7398,
  2020.

\bibitem{DETR}
N.~Carion, F.~Massa, G.~Synnaeve, N.~Usunier, A.~Kirillov, and S.~Zagoruyko,
  ``End-to-end object detection with transformers,'' in \emph{{ECCV}},
  A.~Vedaldi, H.~Bischof, T.~Brox, and J.~Frahm, Eds., vol. 12346, 2020, pp.
  213--229.

\bibitem{grid}
X.~Lu, B.~Li, Y.~Yue, Q.~Li, and J.~Yan, ``Grid {R-CNN},'' in \emph{CVPR},
  2019, pp. 7363--7372.

\bibitem{UODD}
L.~Jiang, Y.~Wang, Q.~Jia, S.~Xu, Y.~Liu, X.~Fan, H.~Li, R.~Liu, X.~Xue, and
  R.~Wang, ``Underwater species detection using channel sharpening attention,''
  in \emph{ACM MM}, 2021, pp. 4259--4267.

\bibitem{FERNet}
B.~Fan, W.~Chen, Y.~Cong, and J.~Tian, ``Dual refinement underwater object
  detection network,'' in \emph{ECCV}, vol. 12365, 2020, pp. 275--291.

\bibitem{UDD}
C.~Liu, Z.~Wang, S.~Wang, T.~Tang, Y.~Tao, C.~Yang, H.~Li, X.~Liu, and X.~Fan,
  ``A new dataset, poisson gan and aquanet for underwater object grabbing,''
  \emph{IEEE TCSVT}, pp. 1--1, 2021.

\bibitem{watergan}
J.~Li, K.~A. Skinner, R.~M. Eustice, and M.~Johnson{-}Roberson, ``Watergan:
  Unsupervised generative network to enable real-time color correction of
  monocular underwater images,'' \emph{{IEEE} Robotics Autom. Lett.}, vol.~3,
  no.~1, pp. 387--394, 2018.

\bibitem{UEGAN}
C.~Fabbri, M.~J. Islam, and J.~Sattar, ``Enhancing underwater imagery using
  generative adversarial networks,'' in \emph{ICRA}, 2018, pp. 7159--7165.

\bibitem{TSNE}
L.~van~der Maaten and G.~Hinton, ``Visualizing data using t-sne,''
  \emph{Journal of Machine Learning Research}, vol.~9, no.~86, pp. 2579--2605,
  2008.

\bibitem{R-CNN}
R.~B. Girshick, J.~Donahue, T.~Darrell, and J.~Malik, ``Rich feature
  hierarchies for accurate object detection and semantic segmentation,'' in
  \emph{CVPR}, 2014, pp. 580--587.

\bibitem{fast}
R.~B. Girshick, ``Fast {R-CNN},'' in \emph{ICCV}, 2015, pp. 1440--1448.

\bibitem{fasterrcnn}
S.~Ren, K.~He, R.~B. Girshick, and J.~Sun, ``Faster {R-CNN:} towards real-time
  object detection with region proposal networks,'' \emph{IEEE TPAMI}, vol.~39,
  no.~6, pp. 1137--1149, 2017.

\bibitem{SSD}
W.~Liu, D.~Anguelov, D.~Erhan, C.~Szegedy, S.~E. Reed, C.~Fu, and A.~C. Berg,
  ``{SSD:} single shot multibox detector,'' in \emph{ECCV}, vol. 9905, 2016,
  pp. 21--37.

\bibitem{Focal}
T.~Lin, P.~Goyal, R.~B. Girshick, K.~He, and P.~Doll{\'{a}}r, ``Focal loss for
  dense object detection,'' in \emph{ICCV}, 2017, pp. 2999--3007.

\bibitem{free}
X.~Zhang, F.~Wan, C.~Liu, R.~Ji, and Q.~Ye, ``Freeanchor: Learning to match
  anchors for visual object detection,'' in \emph{NIPS}, 2019, pp. 147--155.

\bibitem{yolo}
J.~Redmon, S.~K. Divvala, R.~B. Girshick, and A.~Farhadi, ``You only look once:
  Unified, real-time object detection,'' in \emph{CVPR}, 2016, pp. 779--788.

\bibitem{StS}
K.~Han, A.~Xiao, E.~Wu, J.~Guo, C.~Xu, and Y.~Wang, ``Transformer in
  transformer,'' in \emph{NIPS}, 2021.

\bibitem{PVTv1}
W.~Wang, E.~Xie, X.~Li, D.-P. Fan, K.~Song, D.~Liang, T.~Lu, P.~Luo, and
  L.~Shao, ``Pyramid vision transformer: A versatile backbone for dense
  prediction without convolutions,'' in \emph{ICCV}, 2021, pp. 568--578.

\bibitem{PVTv2}
W.~Wang, E.~Xie, X.~Li, D.-P. Fan, K.~Song, D.~Liang, T.~Lu, P.~Luo, and
  L.~Shao, ``Pvtv2: Improved baselines with pyramid vision transformer,''
  \emph{Computational Visual Media}, vol.~8, no.~3, pp. 1--10, 2022.

\bibitem{UP-DETR}
Z.~Dai, B.~Cai, Y.~Lin, and J.~Chen, ``{UP-DETR:} unsupervised pre-training for
  object detection with transformers,'' in \emph{CVPR}, 2021.

\bibitem{Sea-Thru}
D.~Akkaynak and T.~Treibitz, ``Sea-thru: {A} method for removing water from
  underwater images,'' in \emph{CVPR}, 2019, pp. 1682--1691.

\bibitem{UDCP}
P.~D. Jr., E.~R. do~Nascimento, F.~Moraes, S.~S.~C. Botelho, and M.~F.~M.
  Campos, ``Transmission estimation in underwater single images,'' in
  \emph{{ICCV} Workshops}, 2013, pp. 825--830.

\bibitem{RNGE}
H.~Zhang, Z.~Zha, Y.~Yang, S.~Yan, and T.~Chua, ``Robust (semi) nonnegative
  graph embedding,'' \emph{{IEEE} Trans. Image Process.}, vol.~23, no.~7, pp.
  2996--3012, 2014.

\bibitem{resdual}
M.~Long, H.~Zhu, J.~Wang, and M.~I. Jordan, ``Unsupervised domain adaptation
  with residual transfer networks,'' in \emph{NIPS}, 2016, pp. 136--144.

\bibitem{ResNet}
K.~He, X.~Zhang, S.~Ren, and J.~Sun, ``Deep residual learning for image
  recognition,'' in \emph{{CVPR}}, 2016, pp. 770--778.

\bibitem{ResNetXt}
S.~Xie, R.~B. Girshick, P.~Doll{\'{a}}r, Z.~Tu, and K.~He, ``Aggregated
  residual transformations for deep neural networks,'' in \emph{CVPR}, 2017,
  pp. 5987--5995.

\bibitem{ERH}
H.~Song, L.~Chang, Z.~Chen, and P.~Ren,
  ``Enhancement-registration-homogenization {(ERH):} {A} comprehensive
  underwater visual reconstruction paradigm,'' \emph{{IEEE} Trans. Pattern
  Anal. Mach. Intell.}, vol.~44, no.~10, pp. 6953--6967, 2022.

\bibitem{liuhongwei}
C.~Liu, Z.~Wang, S.~Wang, T.~Tang, Y.~Tao, C.~Yang, H.~Li, X.~Liu, and X.~Fan,
  ``A new dataset, poisson {GAN} and aquanet for underwater object grabbing,''
  \emph{{IEEE} Trans. Circuits Syst. Video Technol.}, vol.~32, no.~5, pp.
  2831--2844, 2022.

\end{thebibliography}

\end{document}